\documentclass[runningheads]{llncs}
\usepackage{graphicx}
\usepackage{subfigure}
\usepackage{CJKutf8}
\begin{document}
\title{A Large-Scale Chinese Short-Text Conversation Dataset}
%
%
\author{Yida Wang\inst{1} \and
Pei Ke\inst{2} \and
Yinhe Zheng\inst{2,3} \and
Kaili Huang\inst{2} \and
Yong Jiang\inst{1} \and
Xiaoyan Zhu\inst{2} \and
Minlie Huang\inst{2} \thanks{Corresponding author}
}

%

\institute{
Tsinghua Shenzhen International Graduate School, Tsinghua University, China
\and
\small{Institute for Artifical Intelligence, State Key Lab of Intelligent Technology and Systems.} 
Beijing National Research Center for Information Science and Technology.
\small{Department of Computer Science and Technology, Tsinghua University, Beijing, China.}
 \and
Samsung Research China - Beijing (SRC-B), Beijing, China. 
\email{\{wangyd18,kp17,hkl16\}@mails.tsinghua.edu.cn},
\email{yh.zheng@samsung.com},
\email{jiangy@sz.tsinghua.edu.cn},
\email{\{zxy-dcs,aihuang\}@tsinghua.edu.cn}
}
\maketitle              

\begin{abstract}
The advancements of neural dialogue generation models show promising results on modeling short-text conversations.
However, training such models usually needs a large-scale high-quality dialogue corpus, which is hard to access.
In this paper, we present a large-scale cleaned Chinese conversation dataset \textit{LCCC}, which contains a base version (6.8 million dialogues) and a large version (12.0 million dialogues).
The quality of our dataset is ensured by a rigorous data cleaning pipeline, which is built based on a set of rules and a classifier that is trained on manually annotated 110K dialogue pairs.
We also release pre-training dialogue models which are trained on LCCC-base and LCCC-large respectively.
The cleaned dataset and the pre-training models will facilitate the research of short-text conversation modeling. All the models and datasets are available at \url{https://github.com/thu-coai/CDial-GPT}.

\end{abstract}

\section{Introduction}
\label{ssec:first}

The success of pre-training models has greatly advanced the research of natural language processing  \cite{qiu2020pretrainingsurvey}. 
While BERT \cite{devlin2018bert} promotes various natural language understanding tasks, GPT \cite{radford2019language} demonstrates state-of-the-art performance on natural language generation. Recently, pre-training models have been applied to dialogue generation tasks and achieved state-of-the art results \cite{zhang2019dialogpt,bao2019plato,adiwardana2020towards}.

In addition to the effective Transformer-based model structures, dialogue corpora also play an important role in the success of open-domain dialogue generation models.
Existing work has adopted massive English dialogue corpora from Twitter \cite{ritter2010unsupervised}, Reddit \cite{mazare2018training}, OpenSubtitles \cite{lison2016opensubtitles2016} and other public resources, which equip the pre-training models with the ability to respond to human in open domains.

However, since there are few large-scale Chinese dialogue corpora, the development of the pre-training models for dialogue generation in Chinese has been hindered. We argue that it's essential to construct high-quality large-scale Chinese dialogue corpora which can further promote pre-training models in Chinese. The main challenge we need to face is the quality control of dialogue data. On the one hand, we cannot rely on crowd-sourcing services because existing work has shown that the amount of data used in the pre-training models is quite large \cite{radford2019language}. On the other hand, if we mainly acquire data from public resources such as social media (Weibo, Twitter, Reddit, etc.), the quality of the data will be hard to ensure.  
As a matter of fact, online social media contains many negative behaviors, including toxic comments (i.e. comments that are rude, disrespectful or otherwise likely to make someone leave a discussion)\footnote{https://www.kaggle.com/c/jigsaw-toxic-comment-classification-challenge}, threats, insults, identity hates, obscene contents, and many more.
These factors can remarkably degrade the generation ability of dialogue generation models, and lead to serious unexpected behaviors, which substantially limit the practical use of dialogue models.

In this work, we construct a large-scale cleaned Chinese conversation dataset called LCCC, which contains two versions, LCCC-base and LCCC-large. LCCC-base is filtered from 79 million conversations crawled from Weibo, while LCCC-large is filtered from the combination of Weibo data and other sources of Chinese corpora. A two-phase pipeline is designed to clean this corpus. Specifically, the first phase utilizes a set of heuristic rules to filter out dialogues with inappropriate content, and the
second phase employs several classifiers that are trained on manually labeled data to filter dialogues further. 
Then we present pre-training models for dialogue generation, 
which are first pre-trained on a Chinese novel dataset and then post-trained on LCCC. 
All the pre-trained models and the datasets are released to facilitate future research.


Our contributions can be summarized as below:
\begin{itemize}
\item We build a large-scale cleaned Chinese conversation dataset called LCCC. It can serve as a benchmark for the study of open-domain conversation generation in Chinese.

\item We present pre-training models for Chinese dialogue generation. Moreover, we conduct experiments to show its performance on Chinese dialogue generation.
Both the models and the data are released for public use.
\end{itemize}

\section{Related work}
\subsubsection{Datasets} We make a brief overview of datasets available for data-driven conversation systems, most of which are constructed based on public resources or crowd-sourcing.

As an important public resource of dialogue corpora, movie scripts such as OpenSubtitles have been used to construct dialogue datasets \cite{lison2016opensubtitles2016,serban2015hierarchical}
. The dialogue content usually depends on the scenes of movies. Another public resource is social media, where Twitter \cite{li2016persona,sordoni2015neural}, Reddit \cite{mazare2018training}, Weibo \cite{wang2013dataset,shang2015neural}, and technical forums \cite{lowe2015ubuntu} have been adopted to build open-domain dialogue datasets. These datasets crawled from public resources are usually at a large scale because the corpora of public resources is abundant. However, they also contain much noise which need to be carefully cleaned.

Other researchers resort to construct high-quality dialogue datesets by crowd-sourcing. These datasets are built for advanced dialogue tasks, such as wizard of wikipedia (WoW) \cite{dinan2018wizard} / document grounded conversations (DoG) \cite{zhou2018dataset} for knowledge grounded dialogue generation, PERSONA-CHAT \cite{zhang2018personalizing} for persona enhanced dialogue generation, and DailyDialog \cite{li2017dailydialog} for emotional conversation generation. These datasets are commonly cleaner than those acquired from public resources, but the amount of these datasets is rather small.

\subsubsection{Pre-training Models on Dialogue Generation} 
Since GPT \cite{radford2018improving} achieves state-of-the-art performance on various text generation tasks, recent work has applied pre-training models to dialogue generation.
DialoGPT \cite{zhang2019dialogpt} presents an English open-domain pre-training model which post-trains GPT-2 \cite{radford2019language} on 147M Reddit conversations. Meena \cite{adiwardana2020towards} trains an Evolved Transformer \cite{so2019evolved} with 2.6B parameters on a massive English social media conversation dataset, which contains 40B words. \cite{li2020empirical} investigates dialogue generation by fine-tuning a Chinese GPT on some small dialogue datasets, where the Chinese GPT is pre-trained on a Chinese corpus mixed with Chinese Wikipedia2 (1.7B words) and Chinese News (9.2B words). 

\section{Datasets}
\label{ssec:three}

We crawled 79M conversations from Weibo. First of all, through a rigorous cleaning process, a cleaned Weibo dataset (LCCC-base) was constructed. 
Then, the 79M conversations were mixed with the several public Chinese conversation datasets, and a larger Chinese conversation dataset (LCCC-large) was obtained via more relaxed cleaning conditions.
The cleaning process includes rule-based and classifier-based filtering.

\subsection{Data Collection} 
\subsubsection{LCCC-base} 
A two-phase data collection scheme is used to construct our raw dialogues. At the first phase, a set of seed users were collected. Specifically, we manually selected a batch of Weibo accounts who follow professional mass media dedicating to publish news. We then regarded the users who post comments under these news as ``high-quality'' users, since robot accounts usually do not pay much attention to these daily news.

At the second phase, we collected dialogues from these seed users. Specifically, the Weibo posts from these users are collected along with the following comments, which are organized in tree structures. Note that any path from a root to a leaf can be regarded as a conversation session. We reconstructed these sessions using a Depth First Search process, and 79 million sessions of raw conversations were constructed. 
We then constructed a cleaned Weibo dataset using the cleaning method described in Section~\ref{ssec:cleaning}.

\subsubsection{LCCC-large}
We collected corpora from multiple open-source repositories, including Chinese Chatterbot Corpus \footnote{https://github.com/gunthercox/chatterbot-corpus}, PTT Gossiping Corpus \footnote{https://github.com/zake7749/Gossiping-Chinese-Corpus}, Subtitle Corpus and Xiaohuangji Corpus \footnote{https://github.com/skdjfla/dgk\_lost\_conv}. These datasets, together with Qingyun Corpus and Tieba Corpus, are cleaned and processed to be single-turn conversational data \footnote{https://github.com/codemayq/chinese\_chatbot\_corpus}. Besides, we collected multi-turn conversational data including Douban Conversation Corpus \footnote{https://github.com/MarkWuNLP/MultiTurnResponseSelection}, E-commerical Conversation Corpus \footnote{https://github.com/cooelf/DeepUtteranceAggregation} and a Chinese chat corpus \footnote{https://github.com/yangjianxin1/GPT2-chitchat}.

We then mixed these datasets with the 79M conversations. 
Using the same cleaning process, but by relaxing the threshold of the classifier described below, we obtained a larger version of our dataset (LCCC-large).
\subsection{Cleaning Process}
\label{ssec:cleaning}   

\begin{table}[t!]
\begin{center}
\begin{tabular}{l|l}
\hline 
\textbf{Type} & \textbf{Case} \\ 
\hline
Platform tag & \tiny \small \begin{CJK}{UTF8}{gbsn}\textbf{回复@精灵小宝贝：}我也失眠了\end{CJK} \\ 
Advertisement & \small \begin{CJK}{UTF8}{gbsn}@张伟丽MMA 前来为\textbf{DW持妆粉底液}实力证言！\end{CJK} \\ 
Generic form & \small \begin{CJK}{UTF8}{gbsn}我也是 我也是 我也是 啊啊 \end{CJK}\\ 
Dirty word & \small \begin{CJK}{UTF8}{gbsn}被\textbf{小婊砸}作的现在满身负能量\end{CJK} \\ 
Special word & \small \begin{CJK}{UTF8}{gbsn}可以试试\textbf{左氧氟沙星}\end{CJK} \\ 
Name & \small \begin{CJK}{UTF8}{gbsn}\textbf{陈绍龙}、你无朋友架啦 \end{CJK} \\ 
Symbol &  \small \begin{CJK}{UTF8}{gbsn}\#( ° \_ ° )\# \end{CJK}\\ 
Platform sign & \small \begin{CJK}{UTF8}{gbsn}文科574报哪里好?最好有师范英语\textbf{【微信】} \end{CJK} \\ 
\hline
Not fluent & Q: \small \begin{CJK}{UTF8}{gbsn}昨晚失眠了 A: 发恶梦扎醒,又发我最怕个蒋尸梦!\end{CJK} \\
Incomplete information & Q: \small \begin{CJK}{UTF8}{gbsn}江南小镇很美呀 A:\textbf{印象}\end{CJK} \\ 
Time-sensitive & Q: \small \begin{CJK}{UTF8}{gbsn}感觉没啥电视好看了 A: 琅琊榜，\textbf{就要大结局了}\end{CJK} \\ 
External noun & Q: \small \begin{CJK}{UTF8}{gbsn}假期要过了 A: \textbf{春节}一过,好想立刻回到岛城\end{CJK} \\
Irrelevant pairs & Q: \small \begin{CJK}{UTF8}{gbsn}差点吧洗面奶当牙膏 A: 绿色是今年的流行色\end{CJK} \\

\hline
\end{tabular}
\end{center}
\caption{\label{Cases: dirty} Cases of noise in the cleaning process.}
\vskip -0.2in
\end{table}
\subsubsection{Rule-based Noise Filtering} We filter out many types of noise via rules including: 
(1) delete the platform tag in the dialogues, such as ``Reply to @***", ``[dog]";
(2) remove URL strings from the text;
(3) split conversations with more than 30 turns into multiple conversations less than 30 turns \cite{shang2015neural};
(4) only keep one copy of the phrases or words that are repeated more than 6 times in one sentence;
(5) remove the dialogues if the response is too long or too short;
(6) remove the dialogues if the response is identified as an advertisement by the method in \cite{wang2013dataset}; 
(7) remove the dialogues if 90\% of tri-grams in the response are high-frequency tri-grams  \cite{zhang2019dialogpt};
(8) remove the dialogues if the response has some specific forms of generic responses; 
(9) remove the dialogues in which the response is the same as the post.

We also construct blacklists containing the following noise:
(1) dirty words, sensitive words, and dialect;
(2) special topics words such as levofloxacin;
(3) name, appellation and unknown abbreviation;
(4) special symbols and emoji;
(5) platform signs such as ads, pictures, and videos related words.
The dialogue will be removed if it contains words that appear in the blacklist.
Some cases are shown in Tabel~\ref{Cases: dirty} (top).

\subsubsection{Classifier-based Filtering}
In addition to the rule-based method, we also introduce classifier-based filtering.
Many types of noise in terms of semantics and grammar, and some context-dependent conversations\footnote{The understanding of the first post depends on other context beyond the post.}
are hard to be filtered with rules. 
So we built two BERT classifiers for more elaborate filtering. 
We evaluated precision, recall, and F-score with different confidence scores to choose the best confidence threshold.

The first BERT classifier was trained on manually labeled 100,000 conversations. A dialogue is labeled noisy if it has the above noise or following noise:
(1) The response is not fluent or there are serious typos in the sentence, (2) The information of the response is incomplete; (3) The topic of dialogue is time-sensitive, (4) Festivals, places, gender and time which are not mentioned in the post appear in the response (5) The post and the response are irrelevant.
Some cases are shown in Tabel~\ref{Cases: dirty} (bottom).
The classification accuracy reaches 73.76\% on the test set.

In social media, many conversations inevitably depend on external contexts beyond the text \cite{serban2015survey,wang2013dataset} making them hard to understand \cite{li2016detecting}. 
To alleviate this problem, the second BERT classifier was trained on a manually labeled dataset containing 10,000 utterances.
The classification accuracy reaches 77.60\% on the test set.

\subsection{Statistics and Results}
%
\begin{table}[t!]
\begin{center}
\begin{tabular}{l | c c | c c }
\hline  & \textbf{Single-Turn} & \textbf{Multi-Turn} & \textbf{Single-Turn} & \textbf{Multi-Turn} \\ 
\hline
\small Raw dialogs & 52,708,955 & 26,749,365 & 63,251,887 & 28,189,952 \\
\small Cleaned dialogs & 3,354,277 & 3,466,278 & 7,273,804 & 4,733,955 \\
\small Utterances & 6,708,554 & 13,365,268 & 14,547,608 & 18,341,167 \\
\small Characters & 68,559,727 & 163,690,614 & 162,301,556 & 217,776,649 \\
\small Vocabulary size & 372,063 & 666,931 & 662,514 & 690,027 \\
\small Avg. words & 6.79 & 8.32 & 7.45 & 8.14 \\
\small Avg. turns & 2  & 3.86 & 2  & 3.87 \\
\hline
\end{tabular}
\end{center}
\caption{\label{Tab:LCCC statistic} Statistics of LCCC-base (left) and LCCC-large (right).}
\vskip -0.2in
\end{table}
\begin{table}[t!]
\begin{center}
\begin{tabular}{l l l l l }
\hline \textbf{Dataset} & \textbf{Corpus Statistics} & \textbf{Source} & \textbf{Topic} & \textbf{Corpus Features} \\ 
\hline
\small DuConv \cite{wu2019proactive} & \begin{tabular}[l]{@{}l@{}} \small 29,858 dialogs\\ \small 9.1 turns per dialog\\ \small 10.6 words per turn\end{tabular} & \begin{tabular}[l]{@{}l@{}} \small Crowd \\ \small source\end{tabular} & \begin{tabular}[l]{@{}l@{}} \small  Films and \\ \small film stars\end{tabular} & \begin{tabular}[l]{@{}l@{}} \small  Knowledge-grounded/\\ \small Proactivity modeling\end{tabular} \\
\hline

\small Douban \cite{wu2016sequential} & \begin{tabular}[l]{@{}l@{}} \small 0.5M dialogs\\ \small 7.69 turns per dialog \\ \small 18.56 words per turn \end{tabular} & \small Douban & \small Open topics & \begin{tabular}[l]{@{}l@{}} \small 0.5M Negative dialogs\end{tabular} \\
\hline

\small\begin{tabular}[l]{@{}l@{}} \small Persona-\\ \small Dialog \cite{zheng2019personalized}\end{tabular}  & \begin{tabular}[l]{@{}l@{}} \small 20.83M dialogs \\ \small 56.26M utterances \\ \small 8.47M user profiles\end{tabular}  & \small Weibo & \small Open topics & \begin{tabular}[l]{@{}l@{}} \small Personalization,rich\\ \small user profiles\end{tabular}  \\
\hline

\small STC \cite{shang2015neural} & \begin{tabular}[l]{@{}l@{}}  \small 4.4M pairs\\ \small 219,905 posts\\ \small 4.3M responses \end{tabular} & \small Weibo & \small Open topics & \begin{tabular}[l]{@{}l@{}} \small One post multiple \\ \small responses\end{tabular} \\
\hline

\small $\mathrm{LCCC_{base}}$ & \begin{tabular}[l]{@{}l@{}} \small 6.8M dialogs\\ \small 2.95 turns per dialog\\ \small 20M utterances \end{tabular}  &  \small Weibo & Open topics &  \begin{tabular}[l]{@{}l@{}} \small Extremely Strict cleaning \\ \small process \end{tabular} \\

\hline
\small $\mathrm{LCCC_{large}}$ & \begin{tabular}[l]{@{}l@{}} \small 12M dialogs\\ \small 2.74 turns per dialog\\ \small 33M utterances \end{tabular}  &  \small Mixup & Open topics &  \small Strict cleaning process \\
\hline
\end{tabular}
\end{center}
\caption{\label{Tab:comp} Comparison between existing Chinese conversation datasets and LCCC.}
\vskip -0.2in
\end{table}
%
%
The statistics about the dataset are shown in Table~\ref{Tab:LCCC statistic}. 
The $Avg. words$ means the average number of words per utterance, and text is tokenized by Jieba  \footnote{https://github.com/fxsjy/jieba}. 
We also estimated the noise level in the STC dataset via our blacklist. Results show that 60\% of conversations in STC contain dirty words, sensitive words, special symbols, etc. The model trained on STC generates five times more blacklist words than that trained on LCCC. In Table~\ref{Tab:comp}, a clear comparison between the existing Chinese dialogue dataset and our dataset is presented.

\section{Models}  
\label{Models}
\subsubsection{Architectures}
The model architecture used in this paper is adopted from GPT \cite{radford2018improving} which is based on transformer  \cite{vaswani2017attention} . The transformer decoder consists of multiple masked multi-head self-attention blocks. In each time step, the self-attention mechanism can only observe the information on the left. 

Given a golden response $y$ = $(y^1, ..., y^L)$ and history utterances of a conversation $U$ = $\{u_0, ..., u_n\}$ where each utterance $u_i=(u^1_i, ..., u^{L_i}_i)$ consists of $L_i$ words, 
our goal is train a generation model via maximum likelihood estimation (MLE)
$P(y|U) = \prod_{j=1}^L P(y^j|y^1, ..., y^{j-1},U)$ to generate 
$u_{n+1}^j$ given $U$ with generated ${u_{n+1}^1,...,u_{n+1}^{j-1}}$
until the whole response is generated with one termination symbol.

\subsubsection{Input Representation}
We concatenate all history utterances into one sequence as a long text described in  \cite{wolf2019transfertransfo}. The input of the model is the sum of word embedding, speaker embedding, and position embedding. The word embedding and position embedding are learned during the pre-training phase, and speaker embedding is learned during the post-training or fine-tuning phase. The speaker embedding is used to indicate the different speakers, and we use the speaker symbol as the separation tokens.  Following BERT \cite{devlin2018bert}, we also add a $[CLS]$ to the start of the sequence and use $[SEP]$ as the end-of-sequence token.

\subsubsection{Training}
Following the work of DialoGPT \cite{zhang2019dialogpt}, our models are post-trained based on a pre-trained Chinese GPT model ($\mathrm{GPT_{Novel}}$) on the conversation dataset we collected. 
For multi-turn dialogue instances, following DialoGPT, We take every sentence in the dialogue from the second sentence to the last sentence as the response of history sentences.
We trained several models on the LCCC-base and LCCC-large, respectively (LCCC-base is cleaner), which are summarized as follows:
\begin{itemize}
    \item $\mathrm{GPT_{Novel}}$ is a 12-layer GPT which is pre-trained for 70 epochs on a Chinese novel dataset consisting of a variety of genres (Comedy, Romance, Mystery) with about 0.5 billion tokens. 
    \item $\mathrm{CDialGPT_{LCCC-base}}$ is a 12-layer GPT which is pre-trained for 70 epochs on the Chinese novel dataset and post-trained for 30 epochs on LCCC-base. 
    \item $\mathrm{CDialGPT2_{LCCC-base}}$ is a 12-layer GPT2 which is pre-trained for 70 epochs on the Chinese novel dataset and post-trained for 30 epochs on LCCC-base. 
    \item $\mathrm{CDialGPT_{LCCC-large}}$ is a 12-layer GPT which is pre-trained for 70 epochs on the Chinese novel dataset and post-trained for 30 epochs on LCCC-large. 
\end{itemize}

All models were optimized by the AdamW  \cite{loshchilov2018fixing} optimizer and the Noam \cite{radford2018improving} learning rate decay method.
The layers of all our models are set to 12, and the number of heads is set to 12.
The dimension of word embedding is set to 768, and the dimension of position embedding is set to 513.
The number of the warmup epoch was set to 1, and the maximum learning rate was 6.25e-5. The batch size was set to 8, and the number of gradient accumulation was set to 64. All the models trained with a vocabulary of 13,088 Chinese characters on four RTX 2080ti GPUs. 
Our implementation is based on the open-source code of Transformers  \footnote{https://github.com/huggingface/transformers} and TransferTransfo  \footnote{https://github.com/huggingface/transfer-learning-conv-ai}

\section{Experiments}
\label{ssec:experiments}

\subsection{Fine-tuning Settings and Implementation Details}
To evaluate the models, we fine-tune all our models mentioned in Section~\ref{Models} on the STC dataset. The STC dataset contains 4.4M conversation pairs. We randomly split the dataset into train, validation, and test sets. The validation set and the test set have 20,000 conversation pairs each without any intersection. 

\subsubsection{Baselines}
First we choose the only Chinese dialogue pre-training model \textbf{GPT2-chitchat}  \footnote{https://github.com/yangjianxin1/GPT2-chitchat} as one baseline which is pre-trained on 500k Chinese conversations based on the GPT2  \cite{radford2019language} model.
Then we choose the the vanilla \textbf{transformer}  \cite{vaswani2017attention} as one of the baselines.
This model is widely applied in machine translation and conversation generation \cite{mazare2018training,zhang2018personalizing}.
We also evaluate the classical sequence-to-sequence \cite{cho2014learning,sutskever2014sequence} model with attention mechanism \cite{bahdanau2014neural} (\textbf{Attn-Seq2Seq}) which has been widely used in conversation generation.

\subsubsection{Training Details} The $\mathrm{GPT_{Novel}}$ is fine-tuned up to 30 epochs. 
All the other models are fine-tuned up to 10 epochs with the same batch size and number of gradient accumulation. The other hyper-parameters are also kept the same as the post-training process.
The transformer and Attn-Seq2Seq are trained until convergence on STC without pre-training. For a fair comparison, Attn-Seq2Seq consists of 6 layers of GRU with Bahdanau \cite{bahdanau2014neural} attention mechanism, and the dimension of hidden states is set to 768. The layers of the transformer are also set to 6. Thus, these two models have comparable parameters with pre-training models. We set the parameters of GPT2-chitchat to be the same with the original repository.
For all the models, we linearly decay the learning rate from 6.25e-5 to 0 using the AdamW optimizer. 

\subsubsection{Case Study}
\begin{table}[t!]

\centering
\subtable[An interactive example of multi-turn dialogue generation.]{
\begin{tabular}{l|l}
\hline 
\textbf{Role} & \textbf{Response} \\ 
\hline
User & \small \begin{CJK}{UTF8}{gbsn}别爱我没结果\end{CJK} \\ 
Bot & \small \begin{CJK}{UTF8}{gbsn}没结果的爱是多么的痛 。\end{CJK} \\ 
User & \small \begin{CJK}{UTF8}{gbsn}但是爱过就可以了\end{CJK} \\ 
Bot & \small \begin{CJK}{UTF8}{gbsn}有时候，爱情就是一场游戏。\end{CJK} \\ 
User & \small \begin{CJK}{UTF8}{gbsn}输的总是我\end{CJK} \\ 
Bot & \small \begin{CJK}{UTF8}{gbsn}我爱你，就像我爱你一样。\end{CJK} \\ 
User & \small \begin{CJK}{UTF8}{gbsn}你可拉到吧\end{CJK} \\ 
Bot & \small \begin{CJK}{UTF8}{gbsn}你可拉到吧。\end{CJK} \\ 
User & \small \begin{CJK}{UTF8}{gbsn}别学我说话\end{CJK}\\ 
Bot & \small \begin{CJK}{UTF8}{gbsn}我说话就是你不行。\end{CJK} \\ 
\hline
\end{tabular}
\label{Cases: interactive}
}
\qquad
\subtable[An example of multi-turn self-playing dialogue generation.]{        
\begin{tabular}{l|l}
\hline 
\textbf{Role} & \textbf{Response} \\ 
\hline
User & \small \begin{CJK}{UTF8}{gbsn}生命的意义是什么\end{CJK} \\ 
Bot1 & \small \begin{CJK}{UTF8}{gbsn}生命的意义是过得比你想象\end{CJK}\\&  \small \begin{CJK}{UTF8}{gbsn}的要好\end{CJK}\\ 
Bot2 & \small \begin{CJK}{UTF8}{gbsn}也许吧\end{CJK} \\ 
Bot1 & \small \begin{CJK}{UTF8}{gbsn}嗯，生命的意义是过得比你\end{CJK}\\& \small \begin{CJK}{UTF8}{gbsn}你想象的要好 \end{CJK}\\ 
Bot2 & \small \begin{CJK}{UTF8}{gbsn}说的好\end{CJK} \\ 
Bot1 & \small \begin{CJK}{UTF8}{gbsn}好什么 \end{CJK}\\ 
Bot2 & \small \begin{CJK}{UTF8}{gbsn}过得比你想象的好\end{CJK}\\ 
Bot1 & \small \begin{CJK}{UTF8}{gbsn}好吧 \end{CJK}\\ 
\hline
\end{tabular}
\label{Cases: self-play}
}
\caption{Two generated examples. In the right table, the user prompt is the initial input (the first row). Both of the bots are the same models ($\mathrm{CDialGPT_{LCCC-large}}$).}
\vskip -0.2in
\end{table}
%
\begin{table}[t!]
\begin{center}
\begin{tabular}{ll}
\hline 
\textbf{Post} & \textbf{Response} \\ 
\hline
\small \begin{CJK}{UTF8}{gbsn}作为一名伟大的吃货你必须知道的事情。\end{CJK} 
 & \small \begin{CJK}{UTF8}{gbsn}大半夜的不睡觉，你是在诱惑我么？\end{CJK}
 \\ 
\small \begin{CJK}{UTF8}{gbsn}惊现特种喵！这身手。可以维护世界和平了吧。\end{CJK}
 & \small \begin{CJK}{UTF8}{gbsn}这猫太有才了，这猫很有想象力啊！\end{CJK}
 \\ 
\small \begin{CJK}{UTF8}{gbsn}超萌的考拉。不怕痒，挠吧。\end{CJK}
 & \small \begin{CJK}{UTF8}{gbsn}我的小心脏要跳出来了\end{CJK}
 \\ 
\small \begin{CJK}{UTF8}{gbsn}你觉得最好看的英剧Top3是什么？\end{CJK}
 & \small \begin{CJK}{UTF8}{gbsn}我的英剧top3是《老婆婆》\end{CJK}
 \\ 
\small \begin{CJK}{UTF8}{gbsn}无论身在何方，我心里满满都是你！\end{CJK}
 & \small \begin{CJK}{UTF8}{gbsn}此时此刻，我只想说一句：好可爱 ！\end{CJK}
 \\ 
\small \begin{CJK}{UTF8}{gbsn}安东尼刷屏，可惜小手太小，就是投不进啊\end{CJK}
 & \small \begin{CJK}{UTF8}{gbsn}这小子太牛了，真想给他一拳。\end{CJK}
 \\ 
\small \begin{CJK}{UTF8}{gbsn}披萨饼盒瞬间变笔记本支架，这个太酷了！\end{CJK}
 & \small \begin{CJK}{UTF8}{gbsn}这个真的可以有。但是不是很实用\end{CJK}
 \\ 
\small \begin{CJK}{UTF8}{gbsn}无论你多么讨厌你的学校，当你离开久了，你\end{CJK}
 & \small \begin{CJK}{UTF8}{gbsn}还是学校好，没有负担，没有压力 \end{CJK}\\ \small \begin{CJK}{UTF8}{gbsn}还是会想念它。\end{CJK}
 \\ 

\hline
\end{tabular}
\end{center}
\caption{\label{Cases: LCCC-large} Cases generated by $\mathrm{CDialGPT_{LCCC-large}}$ fine-tuned on STC.}
\vskip -0.2in
\end{table}



%
%
Some cases generated by $\mathrm{CDialGPT_{LCCC-large}}$ which is fine-tuned on STC are provided in Table~\ref{Cases: LCCC-large}. These samples suggest that the model has the ability to generate informative responses. Following the work of DialoGPT  \cite{zhang2019dialogpt}, we provide an interactive dialogue sample and a self-playing dialogue sample in Tabel~\ref{Cases: interactive} and Tabel~\ref{Cases: self-play}. 
All the above samples are decoded by top-p sampling \cite{holtzman2019curious} with temperature 0.7.

\subsection{Evaluation}
\begin{table*}
\centering
\begin{tabular}{lcccccccc}
\hline
\scriptsize \textbf{Models} & \scriptsize \textbf{Size} & \scriptsize \textbf{PPL} & \scriptsize \textbf{BLEU-2} & \scriptsize \textbf{BLEU-4} & \scriptsize \textbf{Dist-1} & \scriptsize \textbf{Dist-2} & \scriptsize \textbf{Greedy} & \scriptsize \textbf{Embedding} \\ & & & & & & & \scriptsize \textbf{Matching}  & \scriptsize \textbf{Average}\\
\hline

\scriptsize  $\mathrm{Attn-Seq2Seq}$ & \scriptsize $73M$ & \scriptsize $34.2$ & \scriptsize $3.93$ & \scriptsize $0.90$ & \scriptsize $0.0085$ & \scriptsize $0.1191$ & \scriptsize $0.6584$ & \scriptsize $0.8338$ \\

\scriptsize  $\mathrm{Transformer}$ & \scriptsize $113M$ & \scriptsize $22.10$ & \scriptsize $\textbf{6.72}$ & \scriptsize $3.14$ & \scriptsize $0.0088$ & \scriptsize $0.1397$ & \scriptsize $0.6606$ & \scriptsize $0.8355$ \\

\scriptsize  $\mathrm{GPT2-chitchat}$ & \scriptsize $88M$ & \scriptsize $-$  & \scriptsize $2.28$  & \scriptsize $0.54$ & \scriptsize $\textbf{0.0103}$ & \scriptsize $\textbf{0.1625}$ & \scriptsize $0.6154$ & \scriptsize $0.7894$ \\

\scriptsize  $\mathrm{GPT_{Novel}}$ & \scriptsize $104M$ & \scriptsize $21.27$ & \scriptsize $5.96$  & \scriptsize  $2.71$ & \scriptsize $0.0080$ & \scriptsize $0.1172$ & \scriptsize $0.6612$ & \scriptsize $0.8334$ \\

\scriptsize  $\mathrm{CDialGPT_{LCCC-base}}$ & \scriptsize $104M$ & \scriptsize $18.38$ & \scriptsize $6.48$  & \scriptsize $3.08$  & \scriptsize $0.0083$  & \scriptsize $0.1268$ & \scriptsize $0.6621$ & \scriptsize $0.8354$ \\

\scriptsize  $\mathrm{CDialGPT2_{LCCC-base}}$& \scriptsize $104M$ & \scriptsize $22.76$ & \scriptsize $5.69$ & \scriptsize $2.50$ & \scriptsize $0.0077$  & \scriptsize $0.1087$ & \scriptsize $0.6624$ & \scriptsize $0.8346$ \\

\scriptsize  $\mathrm{CDialGPT_{LCCC-large}}$ & \scriptsize $104M$ & \scriptsize $\textbf{18.23}$ & \scriptsize $6.63$ & \scriptsize $\textbf{3.20}$ & \scriptsize $0.0083$ & \scriptsize $0.1271$ & \scriptsize $\textbf{0.6625}$ & \scriptsize $\textbf{0.8363}$ \\
\hline
\end{tabular}
\caption{\label{Automatic results}Automatic results.}
\vskip -0.2in
\end{table*}
%
\subsubsection{Automatic Metric} We first employed BLEU  \cite{papineni2002bleu} and distinct n-grams \cite{li2015diversity} as our automatic metrics. Since BLEU cannot perfectly reflect the quality of generated results \cite{liu2016not}, we adopted Greedy Matching \cite{rus2012optimal} to evaluate the relevance between posts and generated responses at the word level and Embedding Average \cite{liu2016not} at the sentence level. We also present the perplexity \cite{manning1999foundations} of different models except GPT2-chichat which has different vocabularies with others.

As shown in Table~\ref{Automatic results}, the models trained on LCCC achieves the best perplexity (PPL) but worse BLEU-2 scores than Transformer, which is consistent with previous work  \cite{zhang2019dialogpt,li2020empirical}.  \cite{adiwardana2020towards} shows a strong correlation between perplexity and human evaluation, so our model has competitive performance. The models trained on LCCC also outperforms others in Greedy Matching and Embedding Average. The GPT2-chitchat reaches the highest distinct scores but poor generation quality where we attribute it to the small scale of the model. 

\subsubsection{Human Evaluation}
\begin{table}[t!]
\begin{center}
\begin{tabular}{l c c c c c}
\hline \textbf{Models} & \textbf{+2} & \textbf{+1} & \textbf{+0} & \textbf{Score} & \textbf{Kappa} \\ \hline

\small $\mathrm{Seq2Seq}$ & \small $27.1\%$ & \small $21.4\%$ & \small $51.4\%$ & $0.756$ & \small $0.4544$ \\

\small $\mathrm{Transformer}$ & \small $42.4\%$ & \small $23.6\%$ & \small $34.0\%$ & $1.084$ & \small $0.4429$ \\

\small $\mathrm{GPT2-chitchat}$s & \small $24.3\%$ & \small $19,1\%$ & \small $56.6\%$ & $0.677$ & \small $0.3941$ \\

\small $\mathrm{CDialGPT_{LCCC-base}}$ & \small $46.7\%$ & \small $21.8\%$ & \small $31.5\%$ & $1.152$ & \small $0.3954$ \\

\small $\mathrm{CDialGPT_{LCCC-large}}$ & \small $\textbf{48.6\%}$ & \small $\textbf{24.5\%}$ & \small $\textbf{27.0\%}$ & $\textbf{1.217}$ & \small $0.4203$ \\
\hline
\end{tabular}
\end{center}
\caption{\label{Tab:human evaluation} Human evaluation.}
\vskip -0.2in
\end{table}
We also employed human evaluation to assess the performance of the generation models. 200 samples of each model (total 1000 samples) are randomly presented to 3 judges to evaluate the following aspects based on a 2/1/0 score schema:
\begin{itemize}
\item \textbf{Fluency and Relevance} If a response is grammatically correct, logically consistent, and relevant to the post, it will get 1. Otherwise, it will get 0.
\item \textbf{Informativeness} If a response is fluent, relevant, and additionally rich in content, it will get 2. 
\end{itemize}

The results are shown in Table~\ref{Tab:human evaluation}. We can see that the models trained on LCCC outperform others. 
Although $\mathrm{CDialGPT_{LCCC-base}}$ is not as good as transformer in automatic metrics, $\mathrm{CDialGPT_{LCCC-base}}$ performs slightly better than $Transformer$ in manual evaluation.
And we computed Fleiss kappa \cite{fleiss1971measuring} to measure the annotation agreement of crowd workers. The values range within 0.39-0.44 indicating fair-moderate agreement.

\section{Conclusion}
This paper presents a large-scale cleaned Chinese conversation dataset that is elaborately cleaned by our filtering pipeline. We provide two versions: one is the base version with 6.8M conversations, and the other is a larger version with 12M conversations. To obtain high-quality dialogue data, we design rule-based and classifier-based filtering procedures. 
We also present pre-training models for Chinese dialogue generation, which is trained on the 12M open-domain conversations. All our pre-training models and the dataset are released for public use.

\section{Acknowledgments}
This work was jointly supported by the National Key R\&D Program of China (Grant No. 2018YFC0830200) and NSFC projects (Key project with No. 61936010 and regular project with No. 61876096). We thank THUNUS NExT Joint-Lab for the support.

\bibliographystyle{splncs04}
\bibliography{LCCC}

\end{document}